\documentclass[runningheads]{llncs}
\usepackage[T1]{fontenc}
\usepackage{graphicx}
\usepackage{multirow}
\usepackage{subcaption}
\usepackage{hyperref}
\usepackage{adjustbox}
\usepackage{color}

\begin{document}
\title{Riddle Generation for Learning Resources \thanks{Supported by Gooru (\url{https://gooru.org})} }

%
\author{Niharika Sri Parasa\orcidID{0000-0001-6476-6732} \and
Chaitali Diwan\orcidID{0000-0003-4875-4752} \and
Srinath Srinivasa\orcidID{0000-0001-9588-6550}}
\authorrunning{Parasa.et al.}
%

\institute{International Institute of Information Technology, Bangalore, India \\
\email{\{niharikasri.parasa,chaitali.diwan,sri\}@iiitb.ac.in}
}

\maketitle              
\begin{abstract}
This paper proposes a novel approach to automatically generate conceptual riddles, with the objective of deployment in online learning environments. 
The riddles are generated by creating triples from the learning resources using the BERT language model, which are fed to the k-Nearest Neighbors language model to identify the proximity between properties and their respective contexts. These properties are classified into Topic Markers and Common based on their uniqueness and modeled on an effective instructional strategy called as Concept Attainment Model. Each riddle is passed through the Validator Module that stores all possible answers for the riddles and is used to verify the learner's answers and provide them hints. 
The riddles generated by our model were evaluated by human evaluators and we obtained encouraging results.

\end{abstract}
\section{Introduction and Background}
\label{sec:introductionandbg}
Activity-based learning is achieved by adopting instructional practices that encourage learners to think about what they are learning~\cite{prince2004does}.
One such instructional strategy in pedagogy that is shown to be effective across domains~\cite{kumar2013effect} is the Concept Attainment Model (CAM)~\cite{joyce2003models}.

The CAM promotes learning through a process of \emph{structured inquiry}. This model is designed to lead learners to a concept by requiring them to analyze the examples that contain the attributes of the concept i.e., positive examples, along with the examples that do not contain these attributes i.e., negative examples. An engaging and fun way to present this model to the learners is by structuring the CAM in the form of riddles. 

Although Riddle solving in learning environments motivates
and interests the learner rather than just reading~\cite{doolittle1995using}, most of the previous works~\cite{ritchie2003jape,waller2009evaluating,guerrero2015theriddlerbot,galvan2016riddle} on riddle generation are addressed in the context of computational creativity/humor.
However, apart from the fact that our approach is backed by an effective instructional strategy, it also has an unique methodology of building riddles by identifying and distinguishing semantically closer concepts based on their properties using the pre-trained language model BERT and k-Nearest Neighbor model.

\section{Approach}
\label{sec:approach}
Our proposed method of Riddle generation includes four modules: Triples Creator, Properties Identifier, Generator followed by Validator.
Each learning resource is passed as an input to our Triples Creator module which first extracts noun phrases, adjectives, verbs and phrases comprising of noun and adjectives as attributes/properties associated with a concept. Then the concept and its associated properties are arranged by masking the relation part as follows: concept $<$mask$>$ property. The masked token is then predicted using Bert-Uncased whole word masking language model~\cite{DBLP:journals/corr/abs-1810-04805}
\footnote{ https://huggingface.co/bert-large-uncased-whole-word-masking}.

Consequently a \emph{Lookup Dictionary} is created where keys are concepts and values are the list of triples along with their respective properties.

These triples are fed to the \emph{Properties Identifier} module where the properties are classified into \emph{Topic Markers} and \emph{Common}. \emph{Topic Markers} are the properties that explicitly represent a concept~\cite{rachakonda2014generic} 
and \emph{Common} property is associated with more than one concept.

We use the k-Nearest Neighbor's Language model~\cite{Khandelwal2020Generalization}, which uses a data store and a binary search algorithm KDTree \footnote{https://scikit-learn.org/stable/modules/generated/sklearn.neighbors.KDTree.html} to query the neighbours of the target token given its context. Each triple of a concept 
along with its respective property are passed as queries to the model, returning the distances, neighbours, and their contexts.
If all the contexts relate to the target concept, then the triple is categorized as \emph{Topic Marker}, otherwise, 
it is categorized as \emph{Common}. Subsequently, neighbouring concepts with common properties are extracted for further use.

The Generator module creates riddles through a Optimized mechanism which creates combinations of triples, either of class Topic Marker or Common as positive examples. Riddles generated from \emph{Topic Markers} of a concept are termed as \emph{Easy Riddles} and those from \emph{Common} are termed as \emph{Difficult Riddles}. 

Difficult Riddles accommodate both positive and negative examples of a concept. So, to generate negative examples for those respective positive examples in 2 versions, the module
uses Lookup Dictionary utilizing formerly extracted neighbouring concepts and their properties.

The generated riddles can have one or more answers. So, each riddle is passed through the \emph{Validator} which generates and stores all possible answers to validate learners' answers and provide hints. 

\section{Experiment and Results}
\label{sec:expandresults}
We use a dataset of 200 open learning resources of the zoology domain comprising free-text curated from Wikipedia
\footnote{https://github.com/goldsmith/Wikipedia}. We had 30 human evaluators that are presented with a sample of 20 riddles both easy and difficult along with multiple-choice options and hints, i.e., topic markers.

    %

Our evaluation approach targets to assess the quality of the riddles (syntactic, semantic, and difficulty level), engagement, informativeness and whether they are fit for learning using 3 point Likert scale. It also captures the overall experience of answering riddles using 5 point Likert scale.
For (a) Are the riddles syntactically correct ?, (b)Are the riddles semantically correct ? $\approx$ 70\%-75\% of the evaluators agreed that the generated riddles are semantically and syntactically correct respectively. And (e) Did you find the riddles interesting?, (c) Is the difficulty level of riddles appropriate ?, (f) Would you be willing to learn through riddles ?, $\approx$ 70\% of the evaluators agreed that the riddles are interesting and $\approx$ 60\% agreed on the difficulty level and their adaptability in learning. More than 70\% of the evaluators agreed the experience of answering riddles to be good. 

\section{Conclusions and Future Work}
\label{sec:conclusionandfuture}
We presented a novel approach to automatically generate concept attainment riddles given a representative set of learning resources. The results obtained from our evaluation are encouraging. As part of future work, we plan to use the generated riddles in an e-learning space to understand its efficiency.

\bibliographystyle{splncs04}
\bibliography{references}

\begin{thebibliography}{50}
\bibitem{boumova2008traditional}
V.~Boumov{\'a}.
\newblock Traditional vs. modern teaching methods: Advantages and disadvantages of each.
\newblock PhD thesis, Masarykova univerzita, Filozofick{\'a} fakulta, 2008.

\bibitem{yi2005effective}
J.~Yi.
\newblock Effective ways to foster learning.
\newblock {\em Performance Improvement}, 44(1):34--38, 2005.

\bibitem{joyce2003models}
B.~Joyce, M.~Weil, and E.~Calhoun.
\newblock {\em Models of teaching}.
\newblock 2003.

\bibitem{prince2004does}
M.~Prince.
\newblock Does active learning work? A review of the research.
\newblock {\em Journal of engineering education}, 93(3):223--231, 2004.

\bibitem{orlofsky2001redefining}
D.~D. Orlofsky.
\newblock {\em Redefining Teacher Education: The Theories of Jerome Bruner and the Practice of Training Teachers}.
\newblock ERIC, 2001.

\bibitem{kumar2013effect}
A.~Kumar and M.~Mathur.
\newblock Effect of Concept Attainment Model on Acquisition of Physics Concepts.
\newblock {\em Universal Journal of Educational Research}, 1(3):165--169, 2013.

\bibitem{sukardjo2020effect}
M.~Sukardjo and M.~Salam.
\newblock Effect of Concept Attainment Models and Self-Directed Learning (SDL) on Mathematics Learning Outcomes.
\newblock {\em International Journal of Instruction}, 13(3):275--292, 2020.

\bibitem{habib2019effectiveness}
H.~Habib.
\newblock Effectiveness of Concept Attainment Model of Teaching on Achievement of XII Standard Students in Social Sciences.
\newblock {\em Shanlax International Journal of Education}, 7(3):11--15, 2019.

\bibitem{kalani2009study}
A.~Kalani.
\newblock A Study of the effectiveness of concept attainment model over conventional teaching method for teaching science in relation to achievement and retention.
\newblock {\em International Research Journal}, 2(5):436--437, 2009.

\bibitem{ahmed2012comparative}
I.~Ahmed, A.~A. Gujjar, S.~A. Janjua, and N.~Bajwa.
\newblock A Comparative Study of Effectiveness of Concept Attainment Model and Traditional Method in Teaching of English in Teacher Education Course.
\newblock {\em Language in India}, 12(3), 2012.

\bibitem{haetami2020effect}
A.~Haetami, M.~Maysara, and E.~C. Mandasari.
\newblock The Effect of Concept Attainment Model and Mathematical Logic Intelligence on Introductory Chemistry Learning Outcomes.
\newblock {\em Jurnal Pendidikan dan Pengajaran}, 53(3):244--255, 2020.

\bibitem{doolittle1995using}
J.~H. Doolittle.
\newblock Using riddles and interactive computer games to teach problem-solving skills.
\newblock {\em Teaching of Psychology}, 22(1):33--36, 1995.

\bibitem{denny2000elementary}
R.~A. Denny, R.~Lakshmi, H.~Chitra, and N.~Devi.
\newblock Elementary "Who am I" riddles.
\newblock {\em Journal of Chemical Education}, 77(4):477, 2000.

\bibitem{shaham2013riddle}
H.~Shaham et~al.
\newblock The riddle as a learning and educational tool.
\newblock {\em Creative Education}, 4(06):388, 2013.

\bibitem{okrah2013riddles}
K.~A. Okrah and L.~Asimeng-Boahene.
\newblock Riddles as communicative and pedagogical tool to develop a multi-cultural curriculum in social studies classroom.
\newblock In {\em African Traditional And Oral Literature As Pedagogical Tools In Content Area Classrooms: K12}, page 129. IAP, 2013.

\bibitem{sultan2018implementation}
A.~Z. Sultan, N.~Hamzah, and M.~Rusdi.
\newblock Implementation of Simulation Based-Concept Attainment Method to Increase Interest Learning of Engineering Mechanics Topic.
\newblock In {\em Journal of Physics: Conference Series}, volume 953, page 012026. IOP Publishing, 2018.

\bibitem{devlin-etal-2019-bert}
J.~Devlin, M.-W. Chang, K.~Lee, and K.~Toutanova.
\newblock BERT: Pre-training of Deep Bidirectional Transformers for Language Understanding.
\newblock In {\em NAACL}, pages 4171--4186, 2019.

\bibitem{Khandelwal2020Generalization}
U.~Khandelwal, O.~Levy, D.~Jurafsky, L.~Zettlemoyer, and M.~Lewis.
\newblock Generalization through Memorization: Nearest Neighbor Language Models.
\newblock In {\em ICLR}, 2020.

\bibitem{ritchie2003jape}
G.~Ritchie.
\newblock The JAPE riddle generator: technical specification.
\newblock {\em Institute for Communicating and Collaborative Systems}, 2003.

\bibitem{galvan2016riddle}
P.~Galv{\'a}n, V.~Francisco, R.~Herv{\'a}s, and G.~M{\'e}ndez.
\newblock Riddle generation using word associations.
\newblock In {\em LREC}, pages 2407--2412, 2016.

\bibitem{waller2009evaluating}
A.~Waller, R.~Black, D.~A. O’Mara, H.~Pain, G.~Ritchie, and R.~Manurung.
\newblock Evaluating the standup pun generating software with children with cerebral palsy.
\newblock {\em ACM Transactions on Accessible Computing (TACCESS)}, 1(3):1--27, 2009.

\bibitem{colton2002automated}
S.~Colton.
\newblock Automated puzzle generation.
\newblock In {\em AISB’02 Symposium on AI and Creativity in the Arts and Science}, 2002.

\bibitem{pinter2012automated}
B.~Pint{\'e}r, G.~V{\"o}r{\"o}s, Z.~Szab{\'o}, and A.~L{\H{o}}rincz.
\newblock Automated word puzzle generation using topic models and semantic relatedness measures.
\newblock In {\em Annales Universitatis Scientiarum Budapestinensis}, volume 36, pages 299--322, 2012.

\bibitem{guerrero2015theriddlerbot}
I.~Guerrero, B.~Verhoeven, F.~Barbieri, P.~Martins, and R.~P. y~P{\'e}rez.
\newblock TheRiddlerBot: a next step on the ladder towards computational creativity.
\newblock In {\em International Conference on Computational Creativity}, pages 315--322, 2015.

\bibitem{rachakonda2014generic}
A.~R. Rachakonda, S.~Srinivasa, S.~Kulkarni, and M.~S. Srinivasan.
\newblock A generic framework and methodology for extracting semantics from co-occurrences.
\newblock {\em Data \& Knowledge Engineering}, 92:39--59, 2014.

\end{thebibliography}

\end{document}